\documentclass[letterpaper, 10 pt, conference]{ieeeconf}  

\IEEEoverridecommandlockouts                              

\usepackage{graphics} 
\usepackage{times} 
\usepackage{amsmath} 
\usepackage{amssymb} 
\usepackage{mathtools}
\usepackage{multirow}
\usepackage{algorithm}
\usepackage{algorithmic}
\usepackage{threeparttable}

\usepackage{bbm}
\usepackage{optidef}

\usepackage[nolist]{acronym}

\usepackage{xcolor}

\title{\LARGE \bf
A Data-Driven Distributed Control Scheme: \\
Learning Multi-Objective Agent-Based MPC for Path-Tracking*
}

\author{Jiaming~Zhong$^{1}$,
        Reza~Valiollahi~Mehrizi$^{1}$,
        Yash~Vardhan~Pant$^{2}$,
        and~Amir~Khajepour$^{1}$
\thanks{*This work was supported by Natural Sciences and Engineering Research Council of Canada (NSERC).}
\thanks{$^{1}$Jiaming Zhong, Reza Valiollahi Mehrizi, and Amir Khajepour are with the Mechatronic Vehicle Systems (MVS) Lab, Department of Mechanical and Mechatronics Engineering, University of Waterloo, 200 University Ave West, Waterloo ON, N2L3G1 Canada
        {\tt\small \{j52zhong, rvaliollahimehrizi, a.khajepour\}@uwaterloo.ca}.}%
\thanks{$^{2}$Yash Vardhan Pant is with the Control, Learning, and Logic (CL2) group, Department of Electrical and Computer Engineering, University of Waterloo, 200 University Ave West, Waterloo ON, N2L3G1 Canada
        {\tt\small yash.pant@uwaterloo.ca}.}%
}

\usepackage[hidelinks]{hyperref}
\hypersetup{
  pdftitle={A Data-Driven Distributed Control Scheme: Learning Multi-Objective Agent-Based MPC for Path-Tracking},
  pdfauthor={Jiaming Zhong, Reza Valiollahi Mehrizi, Yash Vardhan Pant, Amir Khajepour},
  pdfsubject={Author accepted manuscript; ITSC 2024; DOI: 10.1109/ITSC58415.2024.10919711}
}
\newcommand{\AAMnotice}{%
  \noindent\begin{minipage}[b]{\textwidth}
  \footnotesize
  \textbf{Author accepted manuscript.}
  Published in the \emph{2024 IEEE 27th International Conference on Intelligent Transportation Systems (ITSC)},
  Edmonton, AB, Canada, 2024, pp.\ 2999--3004.
  The final published version is available at
  \href{https://doi.org/10.1109/ITSC58415.2024.10919711}{doi: 10.1109/ITSC58415.2024.10919711}.
  \par\smallskip
  \copyright~2024 IEEE. Personal use of this material is permitted.
  Permission from IEEE must be obtained for all other uses, in any current or future media,
  including reprinting/republishing this material for advertising or promotional purposes,
  creating new collective works, for resale or redistribution to servers or lists,
  or reuse of any copyrighted component of this work in other works.
  \end{minipage}
}
\newsavebox{\AAMnoticebox}
\newlength{\AAMfooterreserve}
\newlength{\AAMoriginaltextheight}
\makeatletter
\renewcommand{\@pubidpullup}{0pt}
\makeatother

\begin{document}

\sbox{\AAMnoticebox}{\AAMnotice}
\setlength{\AAMfooterreserve}{\dimexpr\ht\AAMnoticebox+\dp\AAMnoticebox+8pt\relax}
\setlength{\AAMoriginaltextheight}{\textheight}
\addtolength{\textheight}{-\AAMfooterreserve}
\pubid{\raisebox{-\AAMfooterreserve}[0pt][0pt]{\usebox{\AAMnoticebox}}}
\AddToHookNext{shipout/after}{\global\textheight=\AAMoriginaltextheight\relax}
\maketitle
\pagestyle{empty}


\begin{abstract}
Agent-based model predictive control (AMPC) has recently been proposed for vehicle systems with various controllers, such as differential braking and torque vectoring, where controllers are regarded as distributed agents contributing to the same objective. 
However, this scheme is challenging in handling multiple conflicting objectives with coupled agents.
A common approach for such tasks is the integrated MPC, where all objectives and agents are stacked together in one optimization. 
Nevertheless, as more agents and objectives are involved, the integrated MPC will face challenges like computational burdens and maintenance difficulties in practice.
To this end, this paper proposes a learning multi-objective AMPC that can improve design flexibility and computing efficiency.
First, under the assumption of information exchange, a multi-objective AMPC tailored from the alternating direction method of multipliers (ADMM) is proposed to decouple the system and achieve the same performance as the integrated scheme iteratively.
Second, a learning-based method for initializing iterations is proposed to accelerate convergence.
In addition, a data management method is proposed for real-time efficiency, and an authentication module is designed for learning reliability.
We compare the proposed scheme against the integrated scheme via a combined path-tracking simulation for autonomous vehicles with various controllers.
The proposed scheme achieves the same control performance as the integrated one while reducing the computational time by 43.5\%. 
Furthermore, the learning-based method saves 88.6\% more computational time than without learning, making it suitable for real-time implementation.

\end{abstract}


\begin{figure}[!t]
  \centering
  \includegraphics[width=3.4in]{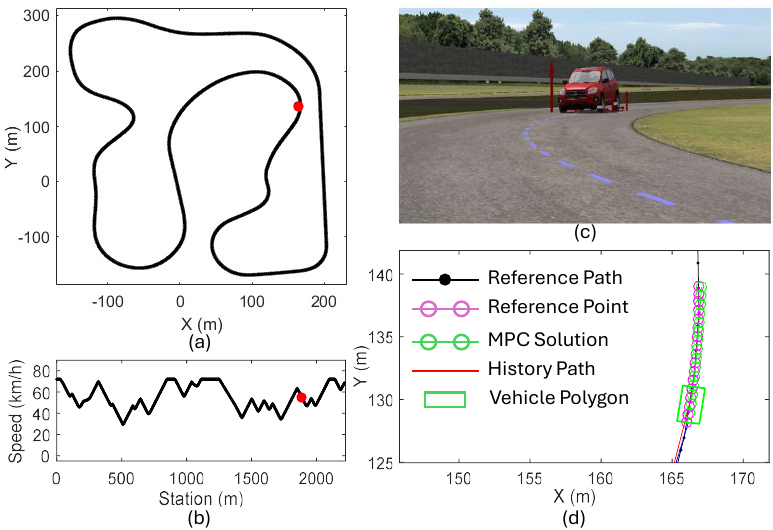}
  \caption{Real-time simulation of the combined path-tracking on a racing track. (a) Global reference path. (b) Reference speed profile. (c) Visualization in CarSim. (d) Control results. (c) and (d) are at station $\approx 1900m$ and marked as red dots in (a) and (b). Details will be given in Section \ref{sec_sim}. A comparison of the converging processes at this position is shown in Fig \ref{fig_sim_B_detail}.}
  \label{fig_sim_setup}
\end{figure}

\section{INTRODUCTION} \label{intro}

\subsection{Motivation}

Model predictive control (MPC) has gained popularity in vehicle control because it optimizes real-time control actions while accounting for constraints. Today's vehicle control systems are increasingly complex as various controllers, such as differential braking and torque vectoring, are incorporated to enhance holistic vehicle performance. A distributed scheme called agent-based MPC (AMPC) has been proposed recently \cite{tang2021agent}, coordinating controllers as agents to achieve the same objective iteratively. However, this scheme is challenging in handling multiple conflicting objectives with coupled agents. For example, path-tracking of autonomous vehicles requires holistic control of longitudinal speed, lateral position, and even stability. The integrated MPC is a common approach for such tasks, where all objectives and agents are stacked together and optimized simultaneously \cite{mazzilli2021integrated}.
Nevertheless, as more agents and objectives are involved, the high-dimensional optimization problem in the integrated MPC will cause computational burdens and maintenance difficulties in practice.
Therefore, the distributed MPC (DMPC), which can decompose objectives into local optimizations, becomes a natural solution for these challenges. 
Some DMPCs \cite{yuan2019mixed}\cite{8802250} adopt a hierarchical architecture to optimize each objective sequentially without any information exchange, which is easy to develop but difficult to obtain the same solution as the integrated one (the global optimum) when objectives are coupled.
To this end, this paper proposes a distributed scheme, the learning multi-objective AMPC, that can achieve the global optimum iteratively through information exchange. Moreover, the learning-based method can improve computing efficiency by reducing convergence iteration.

\subsection{Related Works}

\noindent\textbf{Path-tracking.} Path-tracking of autonomous driving is a comprehensive control that includes longitudinal speed tracking, lateral position tracking, and vehicle stability control. 
Many integrated MPC schemes proposed in the literature for path-tracking have excellent coordination of longitudinal and lateral objectives to achieve global optimization \cite{xiong2022integrated}. 
However, in practice, integrated schemes are not flexible enough in controller design as more electronic control systems that can affect vehicle motion are equipped.

\noindent\textbf{ADMM-based MPC.} The alternating direction method of multipliers (ADMM) uses iterative local optimizations on the augmented Lagrangian equation to approximate the global optimum with solid convergence properties \cite{9632418}.
It has been successfully applied with MPC in applications such as distributed planning and control \cite{zhou2023distributed}\cite{xin2023model}. 
Although ADMM-based MPCs usually have superior computational efficiency to integrated MPCs, their computational time consumption will increase as the number of iterations increases, which depends on the convergence speed and terminal accuracy.
Therefore, we dive into data-driven learning techniques for improving computational efficiency.

\noindent\textbf{Learning-based MPC.} Learning-based MPC (LBMPC) has become one of the main trends in modern controller design.
Gaussian process (GP) is a popular learning tool in combining MPC \cite{hewing2019cautious}. 
Most LBMPC applications are designed to learn the model error or non-linearity, making the MPC predictions more accurate \cite{zhou2022learning}. More recent studies have applied learning to distributed control systems. Authors in \cite{hu2020non} proposed a distributed learning MPC for coupled linear systems, where each subsystem has a learning model that captures the coupling effect of the neighbours. 
Besides, work in \cite{sturz2020distributed} has discussed the recursive feasibility and robustness of learning-based control for distributed schemes.
Inspired by all those fascinating works, this paper proposes a learning scheme for ADMM-based AMPC to improve iterative computational efficiency.

\subsection{Contributions}
The main contributions of this work are as follows:
\begin{itemize}
    \item A distributed control scheme - multi-objective AMPC, tailored from the ADMM, has been proposed. It decouples multi-objective systems into subsystems with information exchange to improve design flexibility. 
    \item A GP-based learning method is developed to predict the contributions for each objective and initialize the iterations to accelerate the convergence process. 
    \item The proposed scheme is applied to a path-tracking task. It achieves the same control performance as the integrated one while reducing the computational time by 43.5\%. Furthermore, the learning-based method saves 88.6\% more computing time than without learning.
\end{itemize}

The paper is structured as follows. The multi-objective control problem and proposed scheme are given in Section \ref{sec_learning_ampc}. Section \ref{sec_path_tracking} demonstrates the path-tracking problem shown in Fig. \ref{fig_sim_setup}. Simulation results are presented with discussions in Section \ref{sec_sim}. Conclusions are summarized in Section \ref{sec_conclusion}.

\section{LEARNING MULTI-OBJECTIVE AMPC} \label{sec_learning_ampc}

\subsection{Integrated MPC}

\begin{figure}[t]
\centering
\includegraphics[scale=0.55]{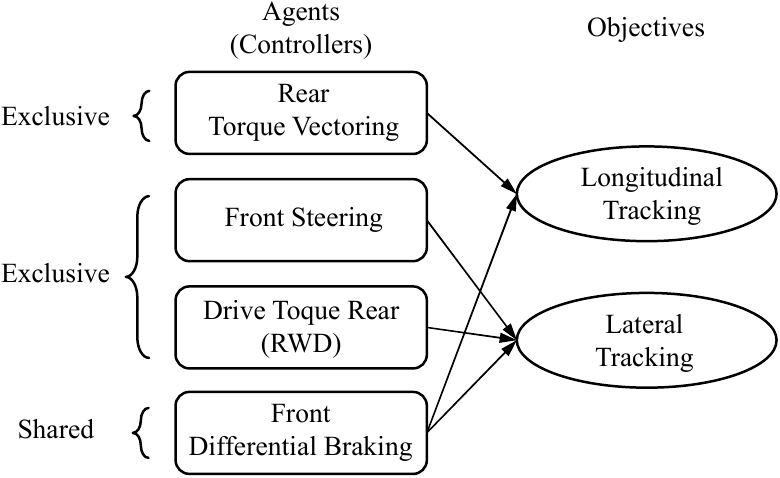}
\caption{Example of the system configuration for the multi-objective path-tracking problem shown in Fig. \ref{fig_sim_setup}. The basic path-tracking problem has two objectives: longitudinal speed tracking and lateral position tracking. These two objectives are coupled through the shared agent (front differential braking). Other agents are exclusive to their objective.}
\setlength{\tabcolsep}{3pt}
\label{fig_example}
\end{figure}

Complex control systems often contain various controllers treated as agents and multiple objectives. Objectives are usually coupled because shared agents affect more than one objective simultaneously, where negotiation between objectives and coordination between agents exist. Path tracking is a typical multi-objective problem with at least longitudinal speed and lateral position tracking tasks completed by multiple agents. An example is shown in Fig. \ref{fig_example}.

Before discussing distributed methods, we establish the integrated MPC \cite{tang2021agent} as a baseline for such systems. This paper uses subscripts $i \in \{1, \ldots, I\}$ for agents and superscripts $j \in \{1, \ldots, J\}$ for objectives. Each $i$-th agent affects the system state $X$ through its control actions $U_i \subseteq U$, where $U = \bigcup_{i} U_{i}$ is the overall set concatenated from agents. Each $j$-th objective only relates to a part of the system state through the output $Y^j$. 
From practical considerations, this paper only focuses on convex linear MPC with quadratic cost functions and linear inequality constraints.
The integrated MPC with the commonly used summation objective can be written as,
\begin{mini}|s|[0]
    {U_i,\dotsc,U_I}
    {J = \sum_{k=1}^{N_p} \Bigl(\sum_{j=1}^{J} h^j(Y^{j}[k]) +  \sum_{i=1}^{I} g_i(U_i)[k-1] \Bigr)} 
    {\label{eq.integMPC}}{}
    \addConstraint{X[k+1] = A_d X[k] + \sum_{i=1}^{I} B_{di} U_i[k] + W_d[k]}
    \addConstraint{Y^{j}[k+1] = C^j X[k+1]}
    \addConstraint{F_i U_i \leq G_i}
\end{mini}
where $N_p$ is the length of the prediction horizon; $k \in \{ 0, \ldots, N_p \}$ is the step in the horizon; $h^j$ and $g_i$ are the quadratic cost functions for the $j$-th objective and the regularization function for the control action $U_i$, respectively; $A_d$, $B_{di}$, $C^j$, and $W_d$ are the discrete system matrix, the $i$-th control matrix, the $j$-th observation matrix, and the disturbance term, respectively; $F_i$ and $G_i$ define polytopic constraints on the inputs $U_i$.
This paper uses $[k]$ with a square bracket for the $k$-th time step and $(q)$ with a round bracket for the $q$-th iteration.
As the number of agents and objectives increases, the matrix dimensions in (\ref{eq.integMPC}) will increase, causing the optimization's computational cost and maintenance complexity to increase.

\subsection{Multi-objective AMPC} \label{sec_ampc}

As a distributed scheme, the proposed multi-objective AMPC tailored from the ADMM consensus formulation \cite{boyd2011distributed} decomposes the system from objectives. Inspired by \cite{ghadimi2014optimal}, a set of global slack variables $Z$ and its local projection $Z^j \subseteq Z$ for the $j$-th objective should be defined to transform the inequality constraints into equality constraints, where $Z = \bigcup_{j} Z^{j}$. Thus, the $j$-th local MPC can be written as,
\begin{mini}|s|[0]
    {U^j}
    {\begin{aligned}J^j=& \sum_{k=1}^{N_p} \Bigl( h^j(Y^{j}[k])) + \displaystyle\sum_{U_i \in U^j} g_i(U_i[k-1]) \Bigr) \\
    &+ L_+(Z^j) \end{aligned}}
    {\label{eq.ConsensusADMM_F3}}{}
    \addConstraint{X[k+1] = A_d X[k] + B_d^j U^{j}[k] + W_d[k]}
    \addConstraint{Y^{j}[k+1] = C^j X[k+1]}
    \addConstraint{F^j U^j - G^j - {Z}^j = 0}
\end{mini}
where $U^j \subseteq U$ and $B_d^j$ are the control actions and the control matrix for the $j$-th objective, respectively; 
$F^i$ and $G^i$ define polytopic constraints on the inputs $U^j$.
$L_+$ is an indicator function that puts an infinite penalty on negative components of the slack variable $Z^j$, making each of the equality constraints in (\ref{eq.ConsensusADMM_F3}) equivalents to the corresponding inequality constraint in (\ref{eq.integMPC}).
Then, the $j$-th local optimization in (\ref{eq.ConsensusADMM_F3}) can be designed as Algorithm \ref{alg:1}.

\begin{algorithm}[ht]
    \caption{Algorithm of Multi-objective AMPC}
    \label{alg:1}
    \begin{algorithmic}[1]
        \STATE \textbf{Input:} matrices for the $j$-th objective
        \STATE \label{step_0} Initialization $\gamma^j = \gamma^{j}(0)$ and $Z^j = Z^{j}(0)$
        \REPEAT
            \STATE \label{step_1} \textbf{Primal optimization} on $U^j$: \\
            $\begin{aligned}
                U^{j}_* = &-{(H^{j} + {A}_{in}^{j^T} \rho^j {A}_{in}^j)}^{-1} \\ 
                & \cdot \Big[ E^{j} + {A}_{in}^{j^T} \rho^{j} 
                (Z^{j} + \bar{\rho^j}^{-1} \gamma^j - b_{in}^j ) \Big]
            \end{aligned} $ 
            \STATE Broadcast $U^j_*$ to all coupled objectives
            \STATE \label{step_2} \textbf{Dual optimization} on $Z^j$:\\
            $\begin{aligned}
                Z_{p*}^j = &\max \Big\{ 0, 
                \displaystyle\frac{1}{m_p} \displaystyle\sum_{j=1}^{m_p} (-{A}_{in}^j U^{j}_* - \bar{\rho^j}^{-1} \gamma^j + b_{in}^j) |_p \Big\} 
            \end{aligned}$ 
            \STATE Copy $Z^j_*$ to update $Z^j$ in all objectives 
            \STATE \label{step_3} \textbf{Lagrange multiplier} $\gamma^j$ update: \\
            $\gamma^j = \gamma^j + \rho^j ({A}_{in}^j U^{j}_* - b_{in}^j + Z^j)$
        \UNTIL{convergence achieved for all objectives}
        \STATE \textbf{Output:} $U^j_*$
    \end{algorithmic}
\end{algorithm}

In Algorithm \ref{alg:1}, $\gamma^j$ is the Lagrange multiplier, and $\rho^j$ is the penalty matrix for convergence.
A $\max\{\}$ operation in the Dual optimization at Step \ref{step_2} handles the positive-only penalty function $L_+$ for constraint satisfaction, where $p$ is the constraint index and $m_p$ is the number of contributed objectives.
$H^{j}$ and $E^{j}$ are the Hessian and linear vector of the local cost function $J^j$ in (\ref{eq.ConsensusADMM_F3}) derived through the batch-formulation along the horizon.
This algorithm's advantage is that it explicitly and efficiently solves the optimization for each objective in its local node.
The proof of the convergence has been discussed in \cite{ghadimi2014optimal} with suggestions on the selection of $\rho^{j}$ leading to the fastest convergence.
The convergence condition of the $j$-th objective is both the primal residual ${\| U^{j}(q)_* - Z^{j}(q) \|}_2^2$ and dual ${\| \rho (Z^{j}(q) - Z^{j}(q-1) )\|}_2^2$ residual are less than certain thresholds.

\subsection{Learning Multi-objective AMPC}

\begin{figure}[t] %
  \centering
  \includegraphics[width=3.3in]{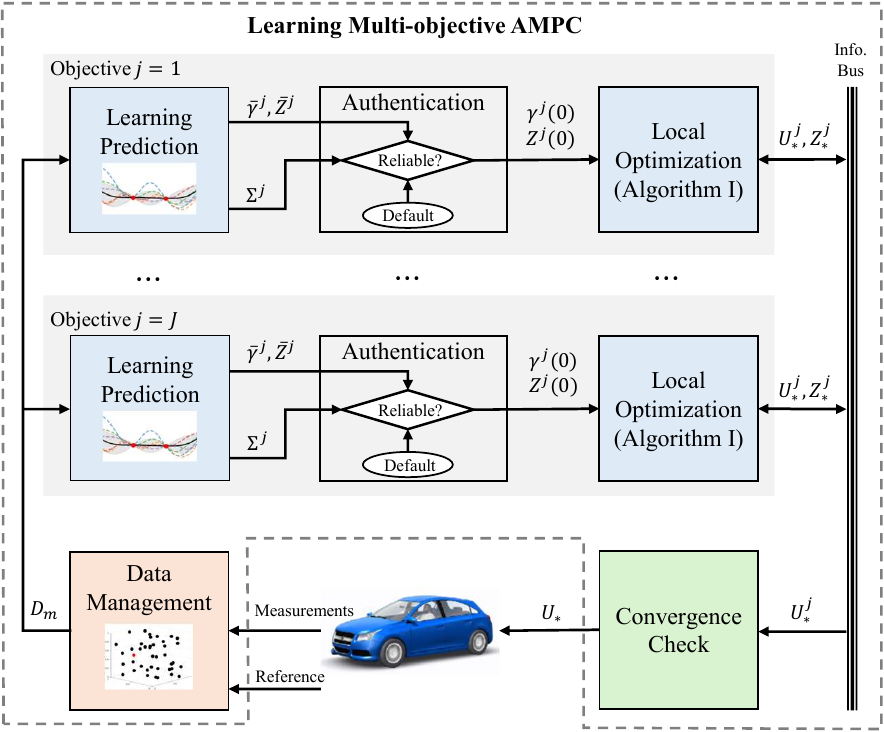}
  \caption{The overall structure of the learning multi-objective AMPC. Each objective has its local optimization module, initialized by a learning prediction module after a reliability authentication. The data management and convergence check can be designed to be either centralized or fully distributed. Local optimizations are based on Algorithm \ref{alg:1}. The convergence check is based on the convergence conditions of all objectives. Learning method and data management will be introduced in Section \ref{sec_learning}.}
  \label{fig_overview}
\end{figure}

In Algorithm \ref{alg:1}, $\gamma^j$ and $Z^j$ will converge to fixed values, indicating the compromise of the $j$-th objective during the convergence process. Therefore, if the values of $\gamma^j$ and $Z^j$ can be predicted from collected data and used as the initial values $\gamma^{j}(0)$ and $Z^{j}(0)$ at Step \ref{step_0} in the Algorithm \ref{alg:1}, the optimized control input $U^j_*$ from the first iteration will be closer to the final global optimum. This way, the number of iterations can be reduced, and the convergence process can be accelerated. Therefore, we propose the learning multi-objective AMPC scheme that leverages a learning-based method to initialize the Algorithm \ref{alg:1}, as shown in Fig. \ref{fig_overview}.

Each $j$-th objective in the scheme has a local optimization based on Algorithm \ref{alg:1}. The optimized control actions $U_*^j$ and slack variables $Z_*^j$ will be updated at each iteration and shared through the information bus. Iterations will end, and the final control action $U_*$ will be output if the convergence check is passed.
Each $j$-th local optimization has a corresponding learning module predicting the mean values of the Lagrange multiplier $\bar{\gamma}^j$ and slack variables $\bar{Z}^j$, indirectly reflecting the compromise during the convergence process. An authentication module will arbitrate based on the prediction variance $\Sigma^j$: Only when the variance $\Sigma^j$ is less than a given threshold, predictions are considered reliable as initial values $\gamma^{j}(0)$ and $Z^{j}(0)$; otherwise, default values (usually are 0 vectors) will be used.

\subsection{GP-based Learning and Data Management} \label{sec_learning}

This work uses GP regression (GPR) as the learning-based method. GPR is a popular non-parametric kernel-based machine learning tool in many engineering applications. Let $\mathcal{D}_n = (\boldsymbol{x}, \boldsymbol{y})$ denotes the dataset with $n$ data points $(x, y)$ where $x$ and $y$ are the input and output vectors for each data point, respectively.
Assume $y = f(x) + \epsilon$ where $\epsilon$ is the additive noises and follow independent, identically distributed Gaussian distributions with zero mean and variances $\sigma_n^2$. GP governs the joint Gaussian distribution of outputs $\boldsymbol{y}$ and the prediction $\overline{y}_*$ for the testing point $(x_*, y_*)$. The predicted mean $\overline{y}_*$ and variance $\Sigma_*$ could be expressed as,
\begin{equation} \label{eq.4}
\begin{split}
    \overline{y}_* &= {K}(x_*, \boldsymbol{x}){[{K}(\boldsymbol{x}, \boldsymbol{x}) + \sigma_n^2 I]}^{-1} \boldsymbol{y} \\
    \Sigma_* &= {K}(x_*, x_*) \\&- 
    {K}(x_*, \boldsymbol{x}){[{K}(\boldsymbol{x}, \boldsymbol{x}) + \sigma_n^2 I]}^{-1} {K}(x_*, x_*)
\end{split} 
\end{equation}
where ${K}(x_*, \boldsymbol{x})$, ${K}(\boldsymbol{x}, \boldsymbol{x})$, and ${K}(x_*, x_*)$ are the covariance matrix whose elements are kernel functions of data pairs.

The online data management process ensures continuous learning ability and high inference efficiency by managing the data density to prevent the data from being too aggregated or clustered. The minimum Euclidian distance is used to determine the action for a new data point:
\begin{enumerate}
    \item \textit{Add the new input point}: The new data point is far from all the existing points in the dataset. 
    \item \textit{Replace the nearest point}: The new data point is close to an existing point, but their outputs are significantly different. New information needs to be updated. 
\end{enumerate}

Furthermore, the K-nearest neighbours (KNN) based on Euclidian distance are used to select a subset $D_m$ for efficient GPR inference.
Then the time complexity will reduce from $\mathcal{O}(n^3)$ to $\mathcal{O}(m^3)$ if only $m\leq n$ data points are selected.

\section{PATH TRACKING PROBLEM} \label{sec_path_tracking}


\begin{figure}[t]
  \centering
  \includegraphics[scale=1.0]{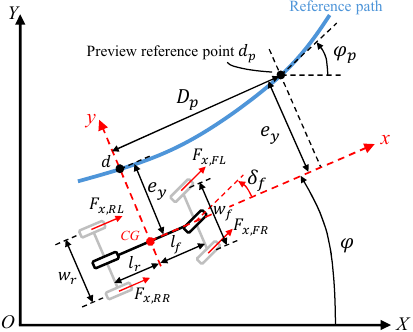}
  \caption{Vehicle and path tracking model.}
  \label{fig_model}
\end{figure}

An extended single-track vehicle model considering the independent left and right torque control and an error-based path-tracking model proposed in \cite{liang2020holistic} are used in this work, as shown in Fig. \ref{fig_model}.
The single-track model is obtained as,
\begin{equation} \label{eq_Model}
\begin{aligned}
&\dot{v}_y = -\frac{c_f+c_r}{m v_x} v_y - (v_x+\frac{l_f c_f - l_r c_r}{m v_x}) r + \frac{c_f \delta_f}{m} \\ 
&\dot{r} = \frac{l_r c_r - l_f c_f}{I_z v_x} v_y - \frac{l^2_f c_f + l^2_r c_r}{I_z v_x} r + \frac{l_f c_f \delta_f + M_{zt}}{I_z}
\end{aligned}
\end{equation}
where $m$ is the vehicle mass; $I_z$ is the yaw inertia; $v_x$, $v_y$ and $r$ are the longitudinal velocity, lateral velocity and yaw rate, respectively; $c_f$ and $c_r$ are the cornering stiffness of the front and rear axles, respectively; $l_f$ and $l_r$ are the distance from the vehicle center of gravity (CG) to front and rear axles, respectively;
$\delta_f$ is the front wheel steering angle, and $M_{zt}$ is the total active yaw moment other control agents generate.
The error-based path-tracking model can be derived as,
\begin{equation} \label{eq.error}
\begin{aligned}
\dot{e}_{y} &= v_x e_{\varphi} + v_y + r D_p \\ 
\dot{e}_{\varphi} &= \dot{\varphi} - \dot{\varphi}_d = r - r_d\\ 
\dot{e}_{u} &= {F_{xt}}/{m} - a_{xd} = \dot{v}_x - \dot{v}_{xd} \\
\end{aligned}
\end{equation}
where $e_y$ is the lateral error from the vehicle CG to the preview point $d_p$ on the reference path; the preview point $d_p$ is calculated based on a preview distance $D_p$ from the closest point $d$ on the reference path; $e_{\varphi}$ is the heading error between the actual heading angle of vehicle $\varphi$ and the desired heading angle $\varphi_d$; $e_{u}$ is the longitudinal speed error between the actual speed $v_x$ and the desired speed $v_{xd}$; $a_{xd}$ is the desired longitudinal acceleration; $F_{xt}$ is the total longitudinal forces. The preview distance $D_p$ is usually selected based on experience and can vary depending on vehicle speed. 

\begin{table}[t]
\caption{System Configuration for Path Tracking}
\label{tb_config}
\centering
\setlength{\tabcolsep}{3.0pt}
\renewcommand{\arraystretch}{1.4}
\begin{threeparttable}
\begin{tabular}{c|c|cc}
\hline\hline
Controller Agent & 
\renewcommand{\arraystretch}{1.1}
\begin{tabular}[c]{@{}c@{}}Control\\ Input\end{tabular} &
\renewcommand{\arraystretch}{1.1}
\begin{tabular}[c]{@{}c@{}}Longitudinal\\ Tracking\end{tabular} & 
\renewcommand{\arraystretch}{1.1}
\begin{tabular}[c]{@{}c@{}}Lateral\\ Tracking\end{tabular}  \\ \hline
\renewcommand{\arraystretch}{1.1}
\begin{tabular}[c]{@{}c@{}}Front Steering\end{tabular} &
$\delta_f$ & 0 & 1  \\ \hline
\renewcommand{\arraystretch}{1.1}
\begin{tabular}[c]{@{}c@{}}Drive Torque Rear\end{tabular} &
$Q_{\text{DTR}}$ & 1 & 0  \\ \hline
\renewcommand{\arraystretch}{1.1}
\begin{tabular}[c]{@{}c@{}}Rear Torque Vectoring\end{tabular} &
$Q_{\text{RTV}}$ & 0 & 1  \\ \hline
\renewcommand{\arraystretch}{1.1}
\begin{tabular}[c]{@{}c@{}}Front Differential Braking\end{tabular} &
\renewcommand{\arraystretch}{1.1}
\begin{tabular}[c]{@{}c@{}}$Q_{\text{FDB,L}}$\\$Q_{\text{FDB,R}}$\end{tabular} & 
1 & 1 \\
\hline\hline
\end{tabular}
\footnotesize
{The Boolean value 1 represents a contribution from the agent to the objective. Otherwise, 0 means no contribution.}
\end{threeparttable}
\end{table}

The $F_{xt}$ and the active yaw moment $M_{zt}$ can be expressed with the small steering angle assumption as,
\begin{equation} \label{eq.fx_mz}
\begin{aligned}
F_{xt} &= {(Q_{\text{FL}} + Q_{\text{FR}} + Q_{\text{RL}} + Q_{\text{RR}})}/{R_w} \\ 
M_{zt} &= {(Q_{\text{FL}} + Q_{\text{FR}}){w_f} + (Q_{\text{RL}} + Q_{\text{RR}}){w_r}}/{2R_w}
\end{aligned}
\end{equation}
where $Q_i, i=\{\text{FL}, \text{FR}, \text{RL}, \text{RR}\}$ are wheel torques at front-left, front-right, rear-left, and rear-right corners, respectively; $w_f$ and $w_r$ are the front and rear track widths; $R_w$ is the wheel radius. 
Wheels torques are generated jointly by various agents. 
The agent configuration is shown in Table \ref{tb_config}, where $\delta_f$, $Q_{\text{DTR}}$, $Q_{\text{RTV}}$, $Q_{\text{FDB, L}}\leq0$, and $Q_{\text{FDB, R}}\leq0$ are steering from front steering (STR) agent, and torque inputs from the drive torque rear (DTR), rear torque vectoring (RTV), and front differential braking (FDB) at left and right wheels, respectively.
Hence, the wheel torques can be expressed as,
\begin{equation} \label{eq.agent_config}
\begin{aligned}
Q_{\text{FL}} &= Q_{\text{FDB,L}} , ~ Q_{\text{FR}} = Q_{\text{FDB,R}}\\ 
Q_{\text{RL}} &= {Q_{\text{DTR}}}/{2} - Q_{\text{RTV}}, Q_{\text{RR}} = {Q_{\text{DTR}}}/{2} + Q_{\text{RTV}} \\ 
\end{aligned}
\end{equation}

Equations (\ref{eq_Model}) $\sim$ (\ref{eq.agent_config}) can be combined into an integrated state-space form as in (\ref{eq.integMPC}), where
\begin{equation} \label{eq.integ_matrix} 
\begin{aligned}
X &= {[e_u, e_y, e_\varphi, v_y, r]}^T, Y ={[e_u, e_y, e_\varphi]}^T\\ 
U &= {[\delta_f, Q_{\text{DTR}}, Q_{\text{RTV}}, Q_{\text{FDB, L}}, Q_{\text{FDB, R}}]}^T
\end{aligned}
\end{equation}

The controller agents can be split into two groups based on Table \ref{tb_config}: longitudinal tracking ($j=\text{long}$) and lateral tracking ($j=\text{lat}$), sharing the FDB agent. For each subsystem, the states, control inputs, and outputs in (\ref{eq.ConsensusADMM_F3}) can be defined as,
\begin{equation} \label{eq.ampc_matrix} 
\begin{aligned}
&U^{\text{long}} = {[Q_{\text{DTR}}, Q_{\text{FDB, L}}, Q_{\text{FDB, R}}]}^T, Y^{\text{long}}={[e_u]}^T\\ 
&U^{\text{lat}} = {[\delta_f, Q_{\text{RTV}}, Q_{\text{FDB, L}}, Q_{\text{FDB, R}}]}^T, Y^{\text{lat}}={[e_y, e_{\varphi}]}^T \\ 
\end{aligned}
\end{equation}


As discussed in Section \ref{sec_learning}, $\gamma^j$ and $Z^j$ from the previous final iteration are collected as data outputs $y$. 
The data inputs $x$ are often empirically selected and filtered through statistical criteria.
According to the system equation, we select the states from measurements and the reference path as inputs. 
Thus, the inputs and outputs of a data point are collected from the previous time step as $x = [e_{u}, e_{y}, e_{\varphi}, v_{y}, r, v_{xd}, \dot{\varphi}_{d}]^T$ and $y^j= [\gamma^j, Z^j]^T$, respectively.

\section{SIMULATION AND RESULTS} \label{sec_sim}

This section presents several comparison simulation results on a racing track scenario in CarSim, with the controller running in Matlab/Simulink, as shown in Fig. \ref{fig_sim_setup}. The integrated MPC was running as a baseline using the \texttt{active-set} algorithm in the Matlab built-in \texttt{quadprog} function.
All the controllers run in real-time at a fixed 20Hz frequency on a laptop with an i7-12700H CPU and 16 GB RAM.

\subsection{Convergence}

\begin{figure}[t]
  \centering
  \includegraphics[width=3.4in]{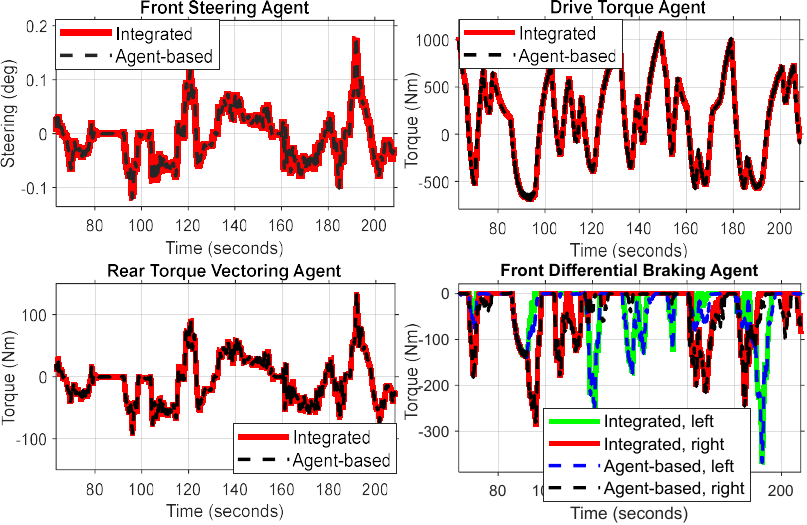}
  \caption{Comparison of control actions between the proposed distributed scheme and the integrated scheme.}
  \label{fig_sim_A}
\end{figure}

\begin{table}[t]
\caption{Quantitative Convergence Accuracy}
\label{tb_err}
\renewcommand{\arraystretch}{1.1}
\begin{threeparttable}
\begin{tabular}{c|ccccc}
\hline\hline
\renewcommand{\arraystretch}{2.0}
\begin{tabular}[c]{@{}c@{}}Control action\end{tabular} &
\renewcommand{\arraystretch}{0.8}
\begin{tabular}[c]{@{}c@{}}STR\\ (deg)\end{tabular} & 
\renewcommand{\arraystretch}{0.8}
\begin{tabular}[c]{@{}c@{}}DTR\\ (Nm)\end{tabular} & 
\renewcommand{\arraystretch}{0.8}
\begin{tabular}[c]{@{}c@{}}RTV\\ (Nm)\end{tabular} & 
\renewcommand{\arraystretch}{0.8}
\begin{tabular}[c]{@{}c@{}}FDB-L\\ (Nm)\end{tabular} & 
\renewcommand{\arraystretch}{0.8}
\begin{tabular}[c]{@{}c@{}}FDB-R\\ (Nm)\end{tabular} \\ \hline
\begin{tabular}[c]{@{}c@{}}Absolute error\tnote{*} \end{tabular} & 0.0001 & 0.9544 & 0.0998 & 3.3464 & 3.4309 \\ \hline\hline
\end{tabular}
\footnotesize
{$^*$The median value of one lap data.}
\end{threeparttable}
\end{table}

First, although \cite{ghadimi2014optimal} theoretically proved the convergence of ADMM-based MPC with constraints, this paper conducts numerical verification through simulations.
The control actions from the proposed multi-objective AMPC were compared with the integrated MPC, as shown in Fig. \ref{fig_sim_A}. 
The median errors between the AMPC and integrated MPC results are very small and acceptable in practice, as shown in Table \ref{tb_err}, showing the convergence of the proposed scheme to the integrated scheme, which generates the global optimum.

\subsection{Improvements by Learning} \label{sim_b}

\begin{figure}[t]
  \centering
  \includegraphics[width=3.4in]{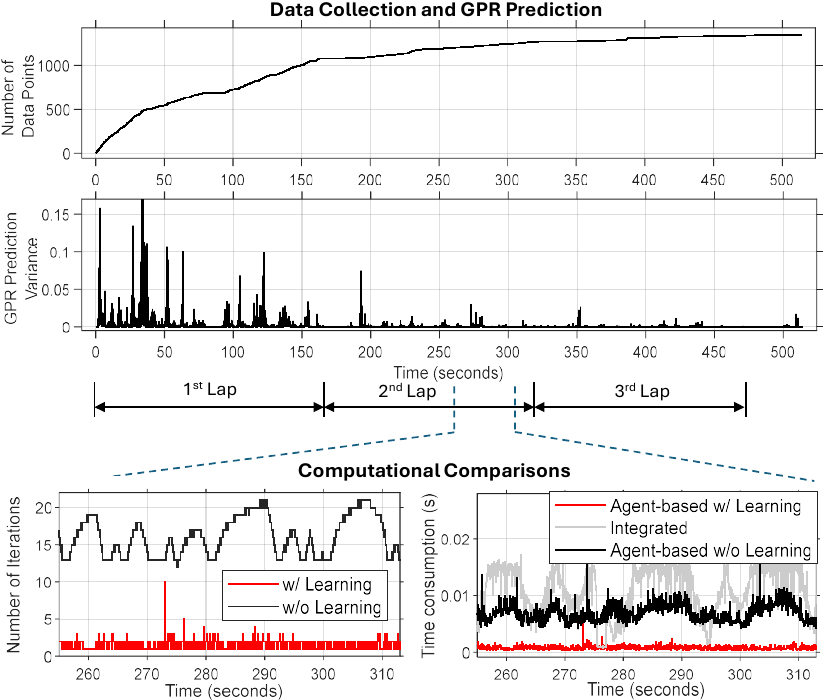}
  \caption{Data collection and learning process. A comparison shows the improvement in computational efficiency.}
  \label{fig_sim_B}
\end{figure}

Several consecutive laps were simulated, starting with an empty dataset, as shown in Fig. \ref{fig_sim_B}. The number of data points grew fast in the first lap but slower after because of the data density management. As more data was collected, the GPR predictions became more accurate as the variance decreased.

The Comparison results on computational efficiency during the second lap are also shown in Fig. \ref{fig_sim_B}. 
The proposed multi-objective AMPC needed about $10 \sim 20$ iterations at each time step to converge, reducing the computational time by 42.5\% compared to the integrated MPC on average.
Furthermore, only less than $5$ iterations were needed if learning the negotiations among objectives. As a result, the learning multi-objective AMPC significantly reduced the computational time by 88.6\% and 92.8\% compared to without learning and the integrated MPC, respectively. 

\begin{figure}[t]
  \centering
  \includegraphics[width=3.4in]{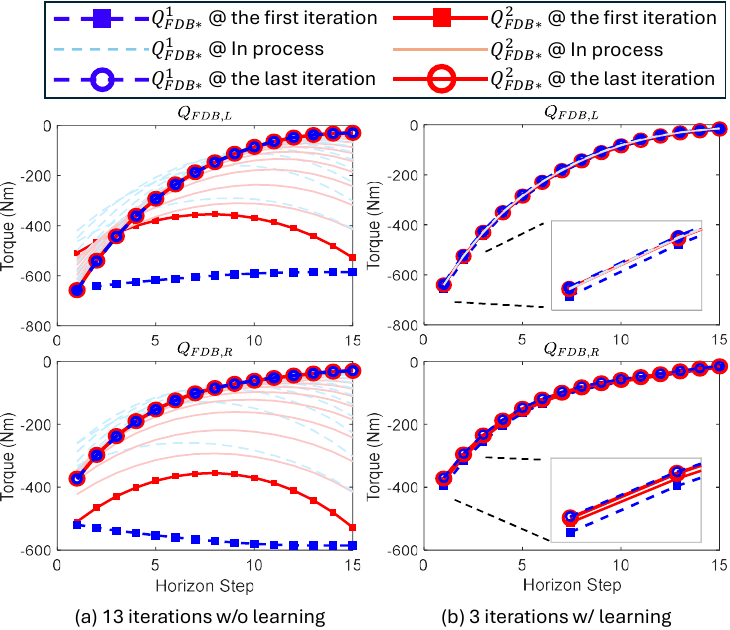}
  \caption{Detailed converging processes of the shared FDB agent at the station $\approx 1900m$ on the racing track, which marked as a red dot in Fig. \ref{fig_sim_setup}.}
  \label{fig_sim_B_detail}
\end{figure}

Detailed converging processes of the shared FDB agent at the station $\approx 1900m$ are compared in Fig. \ref{fig_sim_B_detail}. Results from the first lap without learning predictions are shown in (a) on the left, where FDB torques from the two objectives needed 13 iterations to converge. However, results from the second lap with reliable learned initial values $\gamma^{j}(0)$ and $Z^{j}(0)$ are shown in (b) on the right, where only three iterations were needed. In (b), the control actions $U^{j}_*$ calculated in the first iteration are already very close to the global optimum, significantly reducing the convergence difficulty.

\subsection{Limited Computational Resource}

\begin{figure}[t]
  \centering
  \includegraphics[width=3.4in]{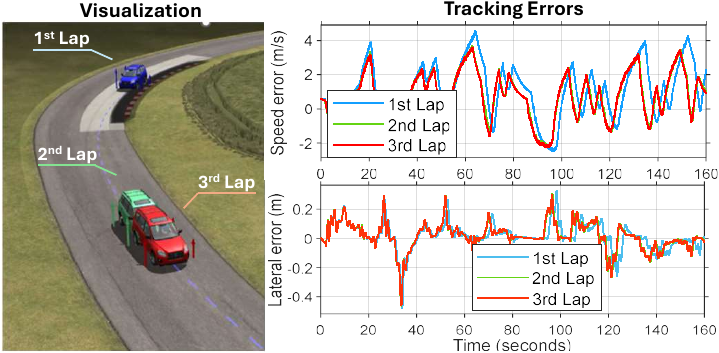}
  \caption{Comparison results of three laps under limited maximum iterations. The speed tracking was improved in the last two laps.}
  \label{fig_sim_C}
\end{figure}

Improvements in computing efficiency can help save computing resource costs. Thus, an interesting assumption is that the maximum number of iterations of AMPC is limited to only 5. Moreover, priorities in the objective are set: lateral tracking is preferred when AMPC cannot converge in the limited number of iterations. 
The comparison simulations of three laps with the same setting as the simulation in Section \ref{sim_b} are conducted, and the results are shown in Fig. \ref{fig_sim_C}.

Since insufficient data was in the first lap, the proposed distributed scheme could not converge within five iterations, so lateral tracking was prioritized. The longitudinal velocity tracking error in the first lap is significantly higher than the last two. The vehicle can neither decelerate in time when entering the corner nor keep up with the acceleration out of the corner. By overlapping the simulation animations together, it's shown that the vehicle in the first lap is much slower than the last two laps.

\section{CONCLUSIONS} \label{sec_conclusion} 

This paper proposes a data-driven distributed control scheme - the learning multi-objective AMPC- tailored from ADMM and can decouple the integrated system into an equivalent but more flexible distributed form. GPR was leveraged to learn the negotiations between objectives for faster convergence. Simulations on a path-tracking task demonstrate its convergence and high computing efficiency.

The proposed scheme is not confined to path-tracking but is promising in many applications as a general data-driven distributed control scheme. It is especially ideal for physically distributed systems with available information exchange, such as interaction-aware planning, multi-vehicle avoidance control, and fleet management. In the future, the proposed method will be expanded to include those interesting topics. Real-time efficiency and convergence performance for systems with a larger number of agents and with stronger non-linearity will be further explored. Also, real-world validation will be implemented.


\addtolength{\textheight}{-12cm}  




\bibliographystyle{bibtex/IEEEtran}
\bibliography{bibtex/IEEEabrv,Paper/ref}

\end{document}